\documentclass[letterpaper]{article} 
\usepackage{aaai19}  
\usepackage{times}  
\usepackage{helvet}  
\usepackage{courier}  
\usepackage{url}  
\usepackage{graphicx}  
\usepackage{lipsum}
\usepackage{amsmath}
\usepackage{amsfonts,amssymb}
\usepackage{caption}
\usepackage{graphicx}
\usepackage{float}
\usepackage[utf8]{inputenc}
\usepackage{booktabs}

\title{Dynamic Transfer Learning for Named Entity Recognition}
\author{ Parminder Bhatia \\
  Amazon \\
  Seattle, USA \\
  \texttt{parmib@amazon.com} \\\And
Kristjan Arumae \\
  Amazon \\
  Seattle, USA \\
  \texttt{arumae@amazon.com} \\\And
  Busra Celikkaya \\
  Amazon \\
  Seattle, USA \\
  \texttt{busrac@amazon.com} \\}
 
\begin{document}

\maketitle

\begin{abstract}
    State-of-the-art named entity recognition (NER) systems have been improving continuously using neural architectures over the past several years.  However, many tasks, including NER, require large sets of annotated data to achieve such performance. In particular, we focus on NER from clinical notes, which is one of the most fundamental and critical problems for medical text analysis.  Our work centers on effectively adapting these neural architectures towards low-resource settings using parameter transfer methods. We complement a standard hierarchical NER model with a general transfer learning framework consisting of parameter sharing between the source and target tasks, and showcase scores significantly above the baseline architecture. These sharing schemes require an exponential search over tied parameter sets to generate the optimal configuration. To mitigate the problem of exhaustively searching for model optimization, we propose the Dynamic Transfer Networks (DTN), a gated architecture which learns the appropriate parameter sharing scheme between source and target datasets. DTN achieves the improvements of the optimized transfer learning framework with just a single training setting, effectively removing the need for exponential search.
\end{abstract}

\section{Introduction}
Natural Language Processing (NLP) applications have been significantly enhanced through advances in neural architecture design.
Tasks such as machine translation, summarization \cite{see2017get}, language modeling \cite{mikolov2010recurrent}, and information extraction have all achieved state of the art systems using deep neural networks, however with a caveat.
These applications require large datasets to generalize well, and naturally sparse domains benefit less from such robust systems.
One such domain is medical data.
Specifically, clinical notes, the free text contents of electronic health records (EHR), have limited availability due to the delicate nature of their content.
Privacy concerns prevent the public release of clinical notes, and furthermore de-identification, and annotation is a lengthy and costly process.

We are interested in Named Entity Recognition (NER)\cite{bhatia2019comprehend} within low-resource areas such as medical domains.
NER is a sequence labeling task similar to part of speech (POS) tagging, and text chunking.
For medical data, NER is an important application as an information extraction tool for downstream tasks such as entity linking \cite{francis2016capturing} and relation extraction \cite{verga2018simultaneously}.
Medical text has challenges that are unique to its domain as well.
Clinicians will often use shorthand or abbreviations to produce patient release notes, with irregular grammar.
This gives the text a significantly less formal grammatical structure than standard NER datasets which often focus on newswire data \cite{RatinovRo09}.
There is also a high degree of variance across sub-domains, which can be attributed to the degree of specialty hospital departments have (e.g. cardiology vs radiology).
While certain medical jargon, and hospital procedure may be invariant of specialty; diseases, treatments, and medications will likely be correlated under these specific sub-domains.
Building an NER system that can learn to generalize well across these is therefore quite difficult, and building individual systems for sub-domains is equally arduous due to the lack of data.
Therefore, we turn towards transfer learning to diminish the effects of data accessibility, and to leverage overlapping representation across sub-domains.

Transfer learning \cite{yang2017transfer} is a learning paradigm that seeks to enhance performance of a target task with knowledge from a source task.
This can take several forms: as pretraining, where a model is first trained for a source task and then some or all weights are used for initialization of the target task; or in place of feature engineering using word embeddings \cite{bhatia2016morphological,bojanowski2016enriching}, a popular approach for most NLP tasks.
We look towards parameter sharing methods \cite{peng2017multi} to transfer overlapped representation from source to target task, when both are NER.

Parameter sharing schemes feature tied weights between layers of a neural network across several tasks.
Finding useful configurations of parameter sharing has been the focus of several recent papers \cite{peng2017multi,yang2016multi,fan2017transfer,guo2018soft,wang2018label}.
As model depth increases the number of possible architectures grows exponentially, and it becomes difficult to exhaustively search through all configurations to choose the best model.
We show that these design choices are a learnable component of the model, and propose a new transfer learning architecture; a generalized neural model which dynamically updates independent and shared components achieving similar scores of models which have been fully tuned.
Our contributions are as follows:
\begin{itemize}
    \item We propose the \textbf{Tunable Transfer Network} (TTN).  A framework which unifies existing parameter sharing techniques into a single model.  This network compartmentalizes all components of our baseline architecture.  Furthermore, we fully explore three degrees of parameter sharing with this system: \textit{hard}, \textit{soft}, and \textit{independent}.  This architecture allows searching for the parameter sharing scheme that best suits the transfer learning setting.
    \item Addressing the large search space problem in TTN, we propose \textbf{Dynamic Transfer Networks} (DTN), a dynamic gated architecture that learns the appropriate parameter sharing between source and target tasks across multiple sharing schemes.  DTN mitigate the issue of exhaustive architecture exploration, while achieving similar performance of the optimized tunable network.
    \item We present a thorough empirical analysis of parameter sharing for low resource named entity recognition on medical data. We also demonstrate DTN's effectiveness on a non-medical dataset achieving best results in such settings.
\end{itemize}

We will first introduce related work as background for NER as well as transfer learning, followed by our proposed architecture, system setup, and dataset information.
We conclude with our findings on low resource settings in both medical and non-medical domains.

\section{Related Work}
NER models achieved their recent success with neural architectures.
In 2016 several works \cite{lample2016neural,chiu2016named,yang2016multi} proposed hierarchical sequence to sequence deep learning frameworks.
The models enjoyed RNN, or CNN encoders, but generally utilized conditional random fields (CRF) as decoders.
Many subsequent works have focused on fine-tuning for speed or parameter size, while keeping this model design at a high level.

Transfer learning for both NER, and other NLP tasks has also been extensively studied.
Here, we will look towards generic models, with more of a focus on those which targeted the medical domain.
\cite{sachan2017effective} leverage unsupervised pretraining in the form of forward and backward language modeling to initialize most of the parameters of an NER architecture. 
Their model was also evaluated on medical data and although the performance increased with pre-training, the evaluation showed low recall from unseen entities.
\cite{yang2016multi} were among the first to explore parameter sharing with the general neural NER architecture.
The authors explored training for NER with other sequence tagging tasks, across multiple languages.
Continuing their work they also correlated task similarity with the number of shared layers in a model \cite{yang2017transfer}.
These architectures generally employed a forked design where feature representations were shared at the lower layers (closer towards the input) and split towards multiple outputs.
For example, tasks in the same language, and with similar labels would share a larger number of layers, whereas sequencing in English and Spanish, regardless of the output space may share only the input embeddings.
The approach of sharing lower level layers was also used for semantic parsing \cite{fan2017transfer}, and co training language models \cite{liu2017empower}.
In the latter only a character level encoder was shared between tasks, and highway units control feature transfer to downstream components.
We employ a similar technique by gating features from multiple inputs at the same layer.
Shared label embedding layers have also shown favorable results \cite{augenstein2018multi,fan2017transfer}.
For multiple tasks a single softmax is used with masking for non-task labels.
The shared embeddings better promote label synergy.

Directly sharing parameters has been widely used, however transfer learning schemes have utilized a soft sharing paradigm as well, where model parameters or outputs are constrained to a similar space.
Most similar to our work, \cite{wang2018label} use two types of constraints to promote shared representations of overlapping output distributions, as well as latent representations.
This work minimizes parameter difference of the CRFs which is derived as the Kullback Leibler divergence upper bound minimization of the target task against the source across overlapping labels from both tasks.
Additionally they constrain the model to produce similar latent representations for tokens with the same tag.
This work is also applied towards NER across several medical sub-domains.
Using soft sharing transfer learning for summarization, \cite{guo2018soft} jointly train three generative models.
Their work was also novel to not have the forked design, in that both the input and output layers were independent.
The same authors used a similar architecture with more ablation on sharing for sentence simplification \cite{guo2018dynamic}.

The parameter sharing architectures discussed here all suffer from the need to exhaustively search for the best architecture, if it is even performed.
Our approach mitigates this procedure by allowing the model to decide what form of parameter sharing it should employ at various layers, and is able to do this during a single training session.

Our model also draws inspiration from pointer networks \cite{vinyals2015pointer,see2017get}.
Pointer networks has shown great performance in assisting generative models augment their  output vocabulary distribution with knowledge of the input sequence.
Our work, however, uses this technique to transfer the signal across several parameter sharing components.

\begin{figure*}
    \centering
    \includegraphics[width=0.91\textwidth]{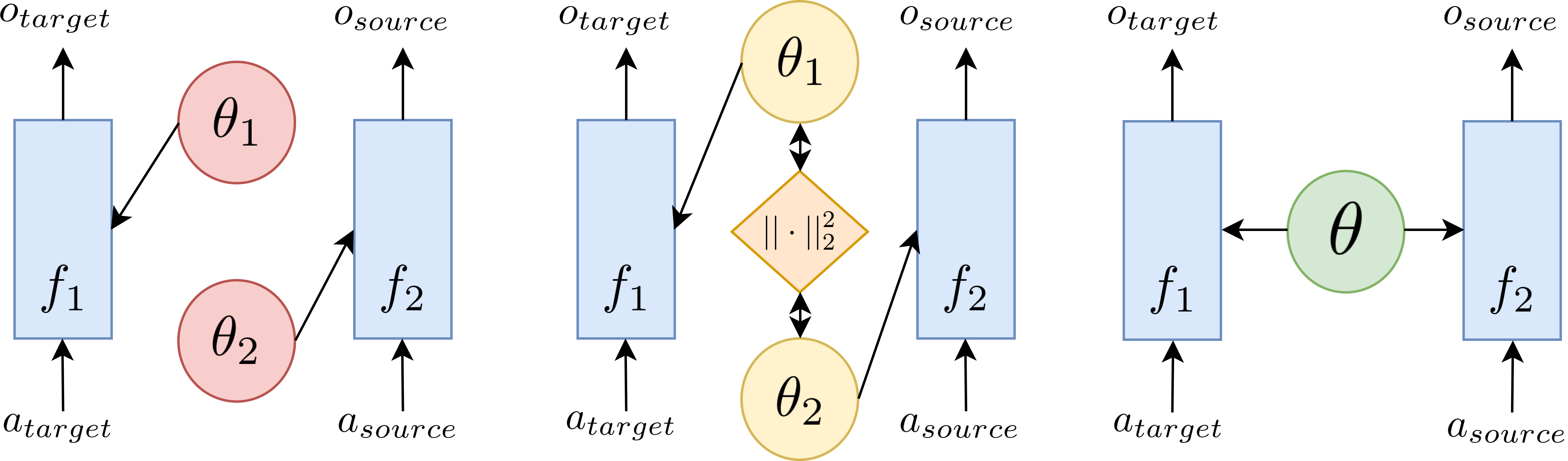}
    \caption{Tunable network architecture: This model is built with the option of independent (left), soft shared (center), or hard shared (right) weights for each of the main components.  The components, presented as $f_1$ and $f_2$, refer to either one of the encoders or the decoder of the target and source task respectively.  Therefore for the tunable NER architecture there are 27 possible configurations.  The blocks in the figure can represent any arbitrary layer in the network, therefore $a$ could refer to input embeddings, or latent representations of tokens, and $o$ will similarly represent any component output.  For both the independent and soft shared approaches $\theta_1$, and $\theta_2$ represent weights assigned to their respective functions, with the center configuration employing the soft sharing constraint $\mathcal{L}_{share}$ between them.}
    \label{fig:TTN}
\end{figure*}
\section{Models}
We first present a standard neural framework for named entity recognition.  
We expand on that architecture by building the Tunable Transfer Network (TTN), to incorporate transfer learning options to each layer.  
Finally, we introduce the Dynamic Transfer Network (DTN), as a trainable transfer learning framework extending the TTN.


\subsection{Named Entity Recognition Architecture}\label{nerarch}
A sequence tagging problem such as NER can be formulated as maximizing the conditional probability distribution over tags $\mathbf{y}$ given an input sequence $\mathbf{x}$, and model parameters $\theta$. 
\begin{equation*}
    P(\mathbf{y} | \mathbf{x}, \theta) = {\displaystyle \prod_{t=1}^{T} P(y_t | x_t, y_{1:t-1}, \theta)}
\end{equation*}
$T$ is the length of the sequence, and $y_{1:t-1}$ are tags for the previous words.
The architecture we use as a foundation is that of \cite{chiu2016named,lample2016neural,yang2016multi}, and while we provide a brief overview of this model we refer the reader to any of these works for architectural insights.
The model consists of three main components: the (i) character and (ii) word encoders, and the (iii) decoder/tagger.

\subsubsection{Encoders}
Given an input sequence $\mathbf{x} \in \mathbb{N}^T$ whose coordinates indicate the words in the input vocabulary, we first encode the character level representation for each word.
For each $x_t$ the corresponding sequence $\mathbf{c}^{(t)} \in \mathbb{R}^{L \times e_c}$ of character embeddings is fed into an encoder.
Here $L$ is the length of a given word and $e_c$ is the size of the character embedding.
The character encoder employs two Long Short Term Memory (LSTM) \cite{hochreiter1997long} units which produce $\overrightarrow{h^{(t)}_{1:l}}$, and $\overleftarrow{h^{(t)}_{1:l}}$, the forward and backward hidden representations respectively, where $l$ is the last timestep in both sequences.
We concatenate the last timestep of each of these as the final encoded representation, $h_c^{(t)} = [\overrightarrow{h^{(t)}_l} || \overleftarrow{h^{(t)}_l}]$, of $x_t$ at the character level.

The output of the character encoder is concatenated with a pre-trained word embedding \cite{pennington2014glove}, $m_t = [h_c^{(t)} || \text{emb}_{word}(x_t)]$, which is used as the input to the word level encoder.
Using learned character embeddings alongside word embeddings has shown to be useful for learning word level morphology, as well as mitigating loss of representation for out-of-vocabulary words.
Similar to the character encoder we use a bidirectional LSTM (BiLSTM) \cite{graves2013speech} to encode the sequence at the word level.
The word encoder does not lose resolution, meaning the output at each timestep is the concatenated output of both word LSTMs, $h_t = [\overrightarrow{h_t} || \overleftarrow{h_t}]$.


\subsubsection{Decoder and Tagger}
Finally the concatenated output of the word encoder is used as input to the decoder, along with the label embedding of the previous timestep.  
During training we use teacher forcing \cite{williams1989learning} to provide the gold standard label as part of the input.
\begin{equation*}
    o_t = \text{LSTM}(o_{t-1}, [h_t || \hat{y}_{t-1}])
\end{equation*}
\begin{equation*}
    \hat{y}_t = \text{softmax}(\mathbf{W}o_t + b^s),
\end{equation*}
where $\mathbf{W} \in \mathbb{R}^{d \times n}$, $d$ is the number of hidden units in the decoder LSTM, and $n$ is the number of tags.
The model is trained in an end to end fashion using a standard cross-entropy objective.

In most of the recent NER literature the focus has been on optimizing accuracy and speed by investigating different neural mechanisms for the three components \cite{strubell2017fast,yang2016multi}.
Both convolutional and recurrent networks have been explored for the encoders, with either conditional random fields (CRF), or single directional RNNs employed as the decoder/tagger.
Since extensive work has been done on this front we fix the design settings and focus only transfer learning while using this common NER architecture.
As described, we use BiLSTM units for both encoders, and a uni-directional LSTM as the decoder.

\begin{figure}[h!]
    \centering
    \includegraphics[scale=0.03]{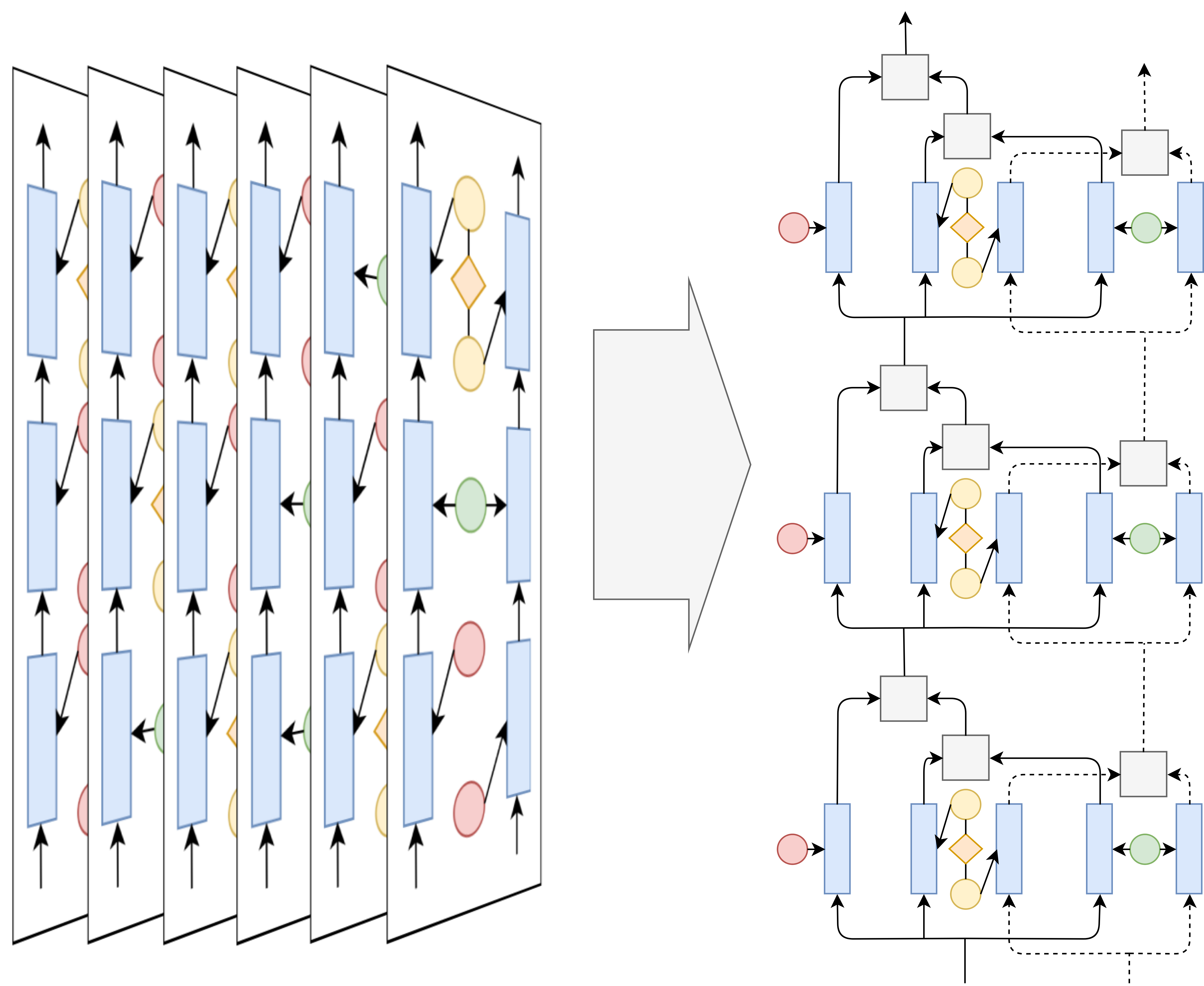}
    \caption{Tunable to Dynamic Transfer Network.  The NER architecture, using all combinations of components from Figure \ref{fig:TTN}, gives us 27 possible architectures (left).  We show, through gating multiple sharing paradigms, that the DTN is able to learn how to produce a similar architecture (right).}
    \label{fig:ta}
\end{figure}
\subsection{Tunable Transfer Network}
The tunable transfer network extends to the components from the previous sections.
Here we focus on how best to benefit from transfer learning with respect to each layer.
To reformulate the architecture from this perspective the model will always train on two tasks, henceforth labeled as \textit{source} and \textit{target}.
Model parameters will be decomposed as:
\begin{equation*}
    \theta = \theta_{source} \cup \theta_{target} \cup\theta_{shared}
\end{equation*}
Source and target parameters are updated by training examples from their respective datasets, while shared parameters receive updates from both tasks.
Updates for parameters will depend on the \textit{batch focus}, meaning for a given forward pass of the model a batch  will contain data from either the source or target task.
During training we shuffle the batches among tasks to allow the model to alternate randomly between them.

We now describe the parameter sharing architectures:
\begin{itemize}
    \item Independent parameters, Figure \ref{fig:TTN} (left).  Relative to the component, the network performs no transfer learning across the two parameters.  For some layers the model performs best when no shared knowledge exists.  This is generally applied to input or output layers.
    
    \item Hard parameter sharing, Figure \ref{fig:TTN} (right).  The parameters of both components reference the same set of weights, and each task in turn updates the same weights.
    
    \item Soft parameter sharing, Figure \ref{fig:TTN} (center).  Individual weights are given to both source and target components, however if this sharing paradigm is present in the model we add an additional segment to the objective: 
    \begin{equation*}
        \mathcal{L}_{share}=||\theta_{source} - \theta_{target} ||^2_2
    \end{equation*}
    Here, we minimize the
$l_2$ distance between parameters as a form of regularization.
    Soft sharing loosely couples corresponding parameters to one another while allowing for more freedom than hard sharing, hence allowing different tasks to choose what sections of their parameters space to share.
    
\end{itemize}


The sharing paradigms from TTN intuitively represent the relatedness of the latent representation of the two tasks for a given component.
Since these are tunable hyperparameters of the architecture, we optimize the model by finding the best configuration of sharing.
Optimizing this involves training $O(M^N)$ unique models, where $M$ is the number of sharing schemes, and $N$ the number of tunable layers. Another problem with the current setup is that for some output distributions the target task may already exhibit high confidence in labels, and introducing a sharing scheme may in fact induce a bias towards the source task. 

\subsection{Dynamic Transfer Network}
Searching across different model architectures motivates us to build a model similar to Figure \ref{fig:ta} which is robust enough to overcome an exponential search of model architecture and achieve similar results compared to the tuned TTN model. 
As mentioned above, being able to tune model architecture is costly, and it is preferable to allow the system to learn how much of a representation to exploit from the source task vs. feedback from its own labels.

\begin{figure}[h!]
    \centering
    \includegraphics[scale=0.07]{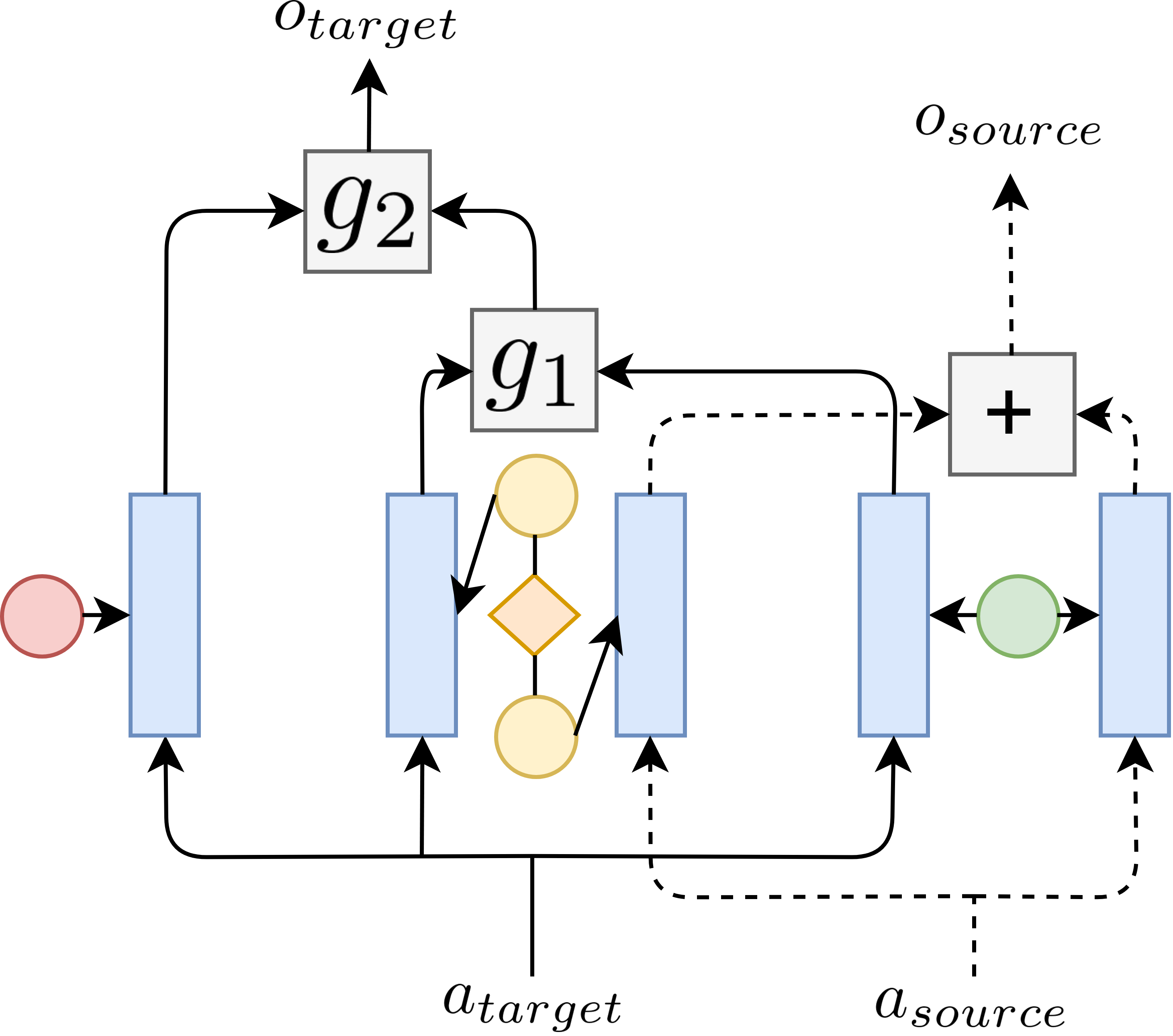}
    \centering
    \caption{Dynamic Transfer Network:  For each encoder and decoder layer of the baseline architecture, we use the DTN architecture.  After passing through their respective RNNs (blue), the target (solid line) uses $g_1$ (Eq. 1) to gate the best representation of the sharing mechanisms.  Similarly, $g_2$ (Eq. 3) gates the output of an independent RNN and $g_1$. The source task (dashed line) has no gating, and is added elmentwise to produce the its respective output.}
    \label{fig:dtn}
\end{figure}

\begin{table*}[!htb]
    \begin{minipage}{.5\linewidth}
      \caption*{(A) Medication (i2b2) to TTP (i2b2) (10\%)}
      \centering
        \begin{tabular}{llll}
            \toprule
            Model & Precision & Recall & $\text{F}_1$\\ 
            \midrule
            Baseline & 55.20 & 48.25 & 51.47\\
            \midrule
            
            \multicolumn{4}{c}{Highest Performance TTN}\\
            \midrule
            \textbf{IIS} & 75.79 & 74.43 & 75.10\\
            
            HIH & 75.65 & 74.29 & 74.96\\
            
            III & 75.42 & 74.34 & 74.87\\
            \midrule
            \multicolumn{4}{c}{Lowest Performance TTN}\\
            \midrule
            
            HSS & 74.92 & 73.71 & 74.31\\
            
            SSI & 75.65 & 72.83 & 74.21\\
            
            SSH & 74.65 & 73.29 & 73.96\\
            
            Avg. & 75.47 & 73.69 & 74.57 $\pm$ 0.24\\
            \midrule
            DTN & 75.65 & 73.61 & 74.46\\
            
            \textbf{DTN (HS)} & 75.83 & 74.09 & 74.95\\
            \bottomrule
        \end{tabular}
    \end{minipage}%
    \begin{minipage}{.5\linewidth}
      \centering
        \caption*{(B) Medication (i2b2) to Medication (Affiliate)  (5\%)}
        \begin{tabular}{llll}
            \toprule
            Model & Precision & Recall & $\text{F}_1$\\ 
            \midrule
            Baseline & 74.37 & 61.49 & 67.29\\
            \midrule
            
            \multicolumn{4}{c}{Highest Performance TTN}\\
            \midrule
            \textbf{SII} & 82.04 & 70.24 & 75.69\\
            
            ISI & 81.69	& 70.37 &	75.60 \\
            
            ISS & 81.75 &	70.21	& 75.54\\
            
            \midrule
            \multicolumn{4}{c}{Lowest Performance TTN}\\
            \midrule
            
            HHH & 80.83 & 69.93 & 74.99\\
            
            SSI & 80.92 &	69.35 &	74.69\\
            
            HII & 81.28 &	69.11 & 74.69\\
        
            Avg. & 81.34 & 70.06 & 75.28 $\pm$ 0.32\\
            \midrule
            
            DTN & 73.78	 & 76.48 & 75.16  \\
            
            \textbf{DTN (HS)} &  80.71	 & 71.11 & 75.60 \\
            \bottomrule
        \end{tabular}
    \end{minipage} 
    \caption{Test set performance during low resource training.  Table 1A displays results from i2b2, transferring from medication to TTP.  Table 1B uses i2b2 medication as source and our affiliate medication data as a target.  The baseline is the current state-of-the art optimized architecture for NER.  For the tunable network (TTN) we indicate the sharing setting alongside each model (S for soft shared, H for hard, and I for independent).  The ordering of the letters follows the that of the components (char enc., word enc., and decoder).  For the sake of space we show only the three best, and worst TTN results, along with the average across all 27 models.  DTN, and DTN (HS) are represented in the bottom two rows.}
\label{table:results}
\end{table*}
Therefore we propose to use the Dynamic Transfer Network (DTN), where gating mechanisms similar to highway units\cite{srivastava2015training}, or pointer generators \cite{see2017get}, control the signal strength from a shared and non-shared component of the network.
We use these gates to choose the best representation between hard and soft sharing, and then between sharing and independent parameters.
This multi-staged gating is similar to the layered pointers used by \cite{mccann2018natural}.

The architecture of DTN is illustrated in Figure \ref{fig:dtn}. To begin, our source and target inputs both pass through their respective RNNs which employ soft (center), and hard (right) sharing, in parallel.
The target and source RNNs take as input $a_{\text{target}}$, and $a_{\text{source}}$ respectively.
This produces two latent representations for both: $h_{\text{t-soft}}$, $h_{\text{s-soft}}$, $h_{\text{t-hard}}$, and $h_{\text{s-hard}}$, where t, and s denote target and source.
We then determine which sharing mechanism was more useful for the target task using a gating function:

\begin{equation}
   g_1 = \sigma(\mathbf{Q}^\intercal h_{\text{t-soft}} + \mathbf{R}^\intercal h_{\text{t-hard}} + \mathbf{S}^\intercal a_{\text{target}} + b_{g_1})
\end{equation}
\begin{equation}
    o_{\text{shared}} = (1 - g_1) h_{\text{t-hard}} + g_1 \cdot h_{\text{t-soft}}
\end{equation}
We also used an independent (left) RNN, to produce a third latent representation for the target, $h_{\text{ind}}$.
Our second gating function takes this, as well as the output of the first gated function as input.
\begin{equation}
   g_2 = \sigma(\mathbf{T}^\intercal h_{\text{ind}} + \mathbf{U}^\intercal o_{\text{shared}} + \mathbf{V}^\intercal a_{\text{target}} + b_{g_2})
\end{equation}
\begin{equation}
    o_{\text{target}} = (1 - g_2) h_{\text{ind}} + g_2 \cdot o_{\text{shared}}
\end{equation}
The final result is a combined representation of the target task as input to subsequent layers.
For both gates, $\sigma$ is the sigmoid function, and $\mathbf{Q}$, $\mathbf{R}$, $\mathbf{S}$, $\mathbf{T}$, $\mathbf{U}$, $\mathbf{V}$, $b_{g_1}$, and $b_{g_2}$ are trainable parameters.
Since our task focuses on how best to adapt the layer towards the target task, the source hidden representations are simply added element-wise to produce: 
\begin{equation*}
    o_{\text{source}} = h_{\text{s-hard}} + h_{\text{s-soft}}
\end{equation*}

The final loss for a network using DTN (Figure \ref{fig:ta}) has the weighted soft sharing regularization objective, along with the cross entropy loss of both tasks.
\begin{equation*}
    \mathcal{L}_{CE} = \mathcal{L}_{\text{target}} + \mathcal{L}_{\text{source}}
\end{equation*}
\begin{equation*}
    \mathcal{L}=\mathcal{L}_{CE}+\lambda \mathcal{L}_{share}
\end{equation*}

TTN has a similar objective, however not all configurations will contain $\mathcal{L}_{share}$.

\subsubsection{Inference}
Both the TTN, and DTN use only parameters for the target task during evaluation and inference.  For example the soft sharing component does not make use of the source parameters at this time.

\section{Experimental Setup}
\subsection{Datasets}

Our work utilizes four main corpora where we employ a tagging scheme that follows an inside, outside, begin, end and singleton (IOBES) format.
We use the public datasets from the 2009 and 2010 i2b2 challenges for medication (Med) \cite{uzuner2010extracting}, and ``test, treatment, problem'' (TTP) entity extraction.
\begin{table}[H]
\centering
\begin{tabular}{llllll}
    \toprule
     & Med & TTP & Affiliate  & CoNLL & Onto \\ 
    \midrule
    Tags & 25 & 13 & 37 & 17 & 73\\
    Notes & 252 & 426 & 1000 & 1393 & 3,637\\
    Tokens & 336K & 416K & 1.5M & 301K & 2M \\
    \bottomrule
\end{tabular}
\caption{Overview of i2b2, affiliate, and newswire datasets.}
\end{table}

The second dataset is obtained through an affiliate, and it is annotated similar to the i2b2 medication challenge.  
Both of the above datasets contain free-text release notes, which have been de-identified.

Additionally, we explore non-medical, newswire data: CoNLL 2003 English \cite{tjong2003introduction} and OntoNotes 5.0 English \cite{pradhan2013towards}.

\subsection{Model Settings}
Word, character and tag embeddings are 100, 25, and 50 dimensions respectively.
Word embeddings are initialized using GloVe \cite{pennington2014glove}, while character and tag embeddings are learned from scratch.
Character, and word encoders have 50, and 100 hidden units respectively.
Decoder LSTM has a hidden size of 50.
Dropout is used after every LSTM, as well as for word embedding input.
We use Adam \cite{kingma2014adam} as an optimizer.
Our model is built using MXNet \cite{chen2015mxnet}.

\begin{table}[!h]
    \caption*{OntoNotes to CoNLL (10\%) }
    \centering
    \begin{tabular}{llll}
        \toprule
        \multicolumn{4}{c}{Highest Performance TTN}\\
        \midrule
        Model & Precision & Recall & $\text{F}_1$\\ 
        \midrule
        HIH & 82.74 & 81.23 & 81.98\\
        SIS & 81.98 & 80.86 & 81.42\\     
        Avg. & 80.66 & 79.47 & 80.06 $\pm$ 1.04\\ 
        \midrule
        \textbf{DTN (HS)} & 82.45 & 81.78 & 82.12\\
        
        \bottomrule
    \end{tabular}

    \caption{CoNLL test set results using 10 \% training data.}
\label{table:results_co}
\end{table}

\subsubsection{DTN Hard-Soft}
We also evaluate a simplified version of the DTN presented in the previous section.
This model, denoted as DTN (HS) under results, learns the best transfer learning setting between soft coupling and hard sharing.
This model retains the first gate (Eq. 1 and 2) from the architecture and uses $o_{\text{shared}}$ as the final target signal for each component.

\subsection{Experiments}
Our models are trained until convergence, and we use the development set of the target task to evaluate performance for early stopping.
We focus on transfer learning in three settings.
The first setting uses only the i2b2 dataset, where the target task is TTP, and the source task is medication.
The second set of experiments uses our affiliate medication data as a target, with i2b2 medication data as the source.
The third task is non-medical, and uses CoNLL 2003 as the target, with OntoNotes 5.0 as the source.
The first and third setting also allows for reproducible performance since the data is publicly available.
We evaluate the performance of our models on $10\%$ of the total target dataset for the first TL setting, and $5\%$ for the second setting.
For non-medical setting, we used $10\%$ of the total target dataset.
The source dataset is not reduced in any of the experiments.
Development and test set are also kept the original size.
The baseline follows the construction of the architecture described in the first section of modeling.

\begin{figure*}
    \includegraphics[width=\textwidth]{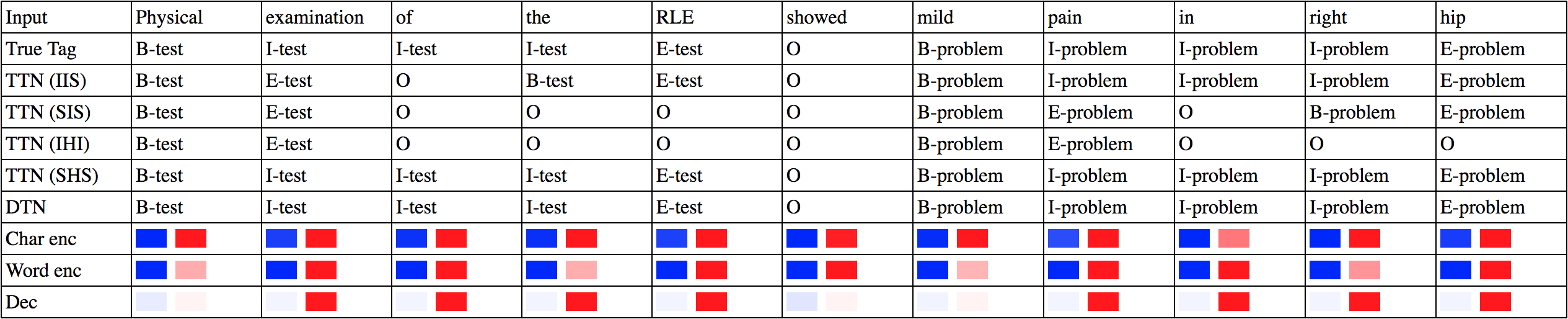}
    \caption{Through visualization we show that the DTN is able to adaptively learn the optimal transfer learning schema across a sequence. To demonstrate this we feed ``Physical examination of the RLE showed mild pain in the right hip'' to the DTN and the four models randomly selected from the top 20  percentile of the best performing TTN models.  We show the ground truth and detected tags for each token, where O denotes any token which is not a named entity.  The bottom three rows indicate the value of the gating signals for the three major components of the DTN. Since each component has two gates (Eq. 1 and 3), we use blue to illustrate $g_1$, and red for $g_2$. A darker color for $g_1$ indicates the model preferred soft sharing over hard.  Appropriately a darker shade of red indicates that the model favored independent over any value of $g_1$. We show the model not only learns how to accurately predict the output tags, but also that it does not follow any specific sharing scheme.}
    \label{fig:samples}
\end{figure*}

\section{Results}
We analyze our results from multiple perspectives.
We first demonstrate the effectiveness of parameter sharing for low resource settings for both experiments in the medical domain.
We also examine model performance across various data percentages to showcase the uniform performance of DTN models.
Furthermore, we explore the gating values across layers to investigate model behavior for the dynamic architecture which suggests why gating can imbibe the characteristics of the best model which varies depending upon the relatedness of the source and target tasks.
We report precision, recall, and macro $\text{F}_1$ on the target data test set.

\subsection{Transfer Learning Performance}
The test set results on all medical data are reported in Table \ref{table:results}.
For the tunable network, we show results for six models (three best, and three worst), as well as the average result across all 27 configurations (three components, and three sharing schemes).

\begin{figure}[H]
    \centering
    \includegraphics[scale=0.5]{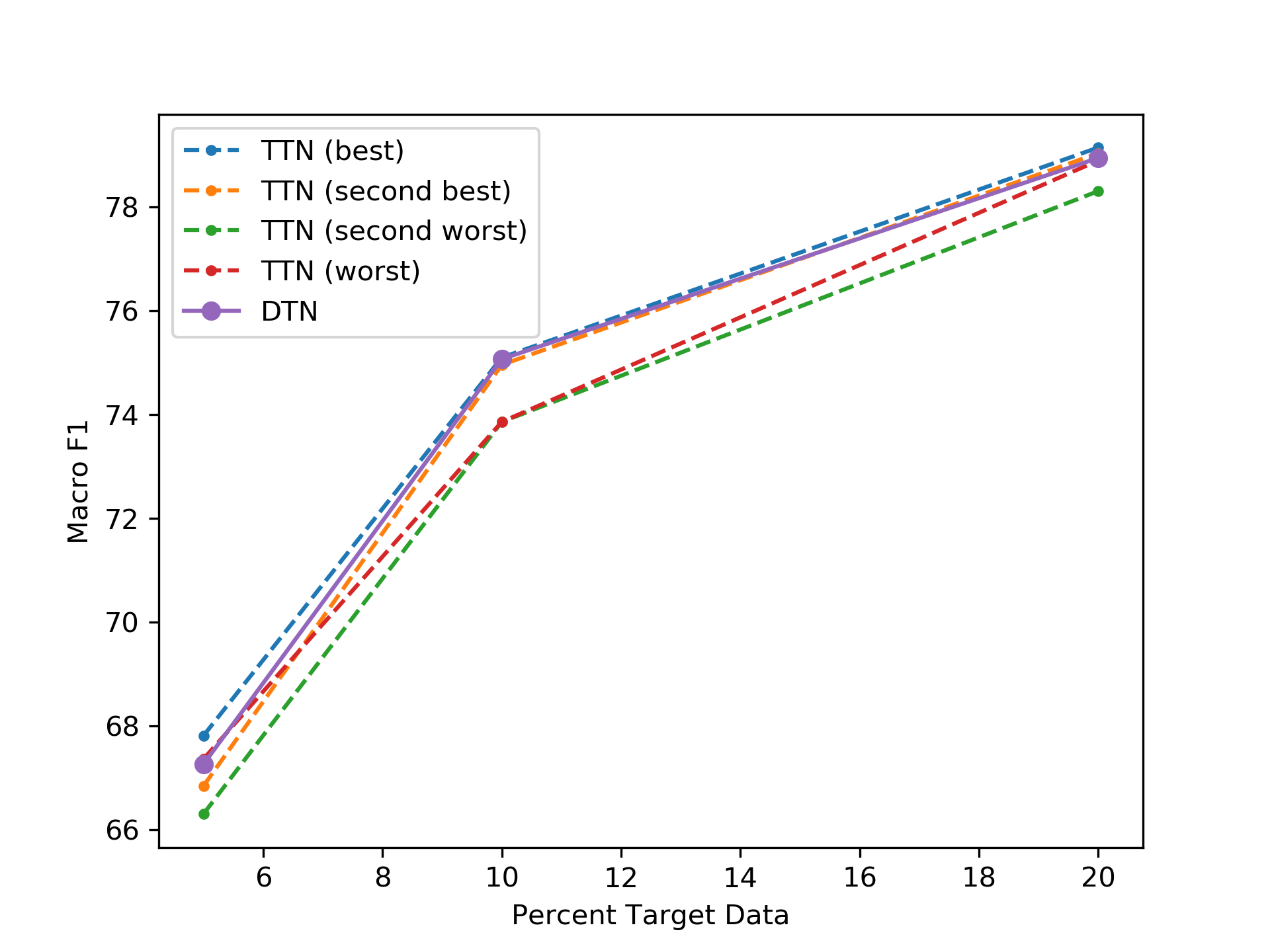}
    \centering
    \caption{We select two best, and worst performing models from the earlier medication to TTP experiments and compare their results against DTN across multiple low resource settings. We observe variance between the best and the worst models and effectiveness of DTN to generalize.}
    \label{fig:ranges}
\end{figure}

For the first setting (Table \ref{table:results}A), there is on average a 36.66$\%$ $\text{F}_1$ gain over the baseline model which indicates that the system greatly benefited from transfer learning.
Similarly there was a 11.21$\%$ increase for TTN across the medication only tasks (Table \ref{table:results}B).
Notably all settings of the tunable model yielded a large margin in performance over both baselines.
More consequential, however, is the range of performance among the tunable models.
We  observed high variance in the first task with the lowest $\text{F}_1$ score (soft-soft-hard) of $73.96$ versus the highest $75.10$ (indep-indep-soft).
The second task had a similar gap of 1 $\text{F}_1$ point between high and low performers.
These results validate the need to search for the best architecture for parameter sharing.

\subsubsection{DTN}
In general DTN performed well, and more intriguing was the capability of DTN (HS), as it always triumphed over its more complex counterpart.
For the first task, the dynamic model achieved a score of $74.46$, and DTN (HS) outperformed all but the best two.
Although slightly less robust in the second TL task, the DTN placed within the $63$'rd percentile of TTN.
DTN (HS) similarly bested all but the top two models.
Model performance is even more commendable for the newswire data.
In Table \ref{table:results_co}, we see a boost over the best performing TTN, with DTN (HS) placing two standard deviations above the TTN average. This validates our hypothesis that the given TTN architectures are not exhaustive to produce the best results and dynamic nature of our model helps in encapsulating the best model setting for a given problem.

Our model provides competitive performance with significantly reduced training time.
Moreover, we show that the DTN is consistent in performance while TTN configurations exhibit variability across training conditions.
Figure \ref{fig:ranges} illustrates this phenomenon.
We chose the best and worst TTN settings for a particular low resource (10\%) setting (from i2b2 medication to TTP) and we see that the rankings are not as tightly coupled when we re-execute the experiments with more (20\%) or fewer (5\%) training samples.
Another example  can be seen from Table \ref{table:results}, where the best performing model, \textbf{SSI}, for medication data sets was one of the worst performing models in case of i2b2 dataset. 
\subsection{Gating}




We take a closer look at output across a sequence in Figure \ref{fig:samples}.
We compare the output of the DTN against better performing TTN models to show how the model adapts when others fail.
The illustration is indicative that the model does not rely on a particular gating scheme consistently.
Instead we observe the changes in gating across a sequence, where the model relies on multiple learning schemes for a given token.



We further analyzed the contributions of DTN between the different sharing schemes.
Upon a closer inspection of the output layer gates as shown in Table \ref{table:avg_gated}, we observe significant variance among parameter sharing across different tag types.
The parameter sharing for tags depends on the relatedness of the target and source tags. 
For example, \textit{Form} is not present in the i2b2 (source) dataset. 
We discern that the decoder sharing scheme for the \textit{Form} tag prefers hard sharing thus smaller value, as it can not leverage much information from the soft sharing scheme.
Overall we observe interesting insights, where a parameter sharing scheme depends on the tag type as well as temporality thereby making RNN more robust to the sensitivity of the data.

\begin{table}[h!]
\centering
\begin{tabular}{llll}
    \toprule
    Component & Char Enc & Word Enc & Decoder \\ 
    \midrule
    Medication Name & 0.64 & 0.91 & 0.77\\
    Form & 0.88 & 0.99 & 0.18\\
    Dosage & 0.69 & 0.99 & 0.26\\
    Frequency & 0.81 & 0.98 & 0.22\\
    \midrule
    Overall & 0.65 & 0.32 & 0.82\\
    \bottomrule
\end{tabular}
\caption{Gate activations are averaged across all tokens from input, for experiment two. These results look at a gate choosing between hard and soft sharing (Eq. 1).  A low value indicates the gate favored hard sharing, whereas a value closer to $1.0$ favors soft sharing. }
\label{table:avg_gated}
\end{table}
\section{Conclusion}
In this paper we have shown that tuning a transfer learning architecture in low resource settings will allow for a more efficient architecture.
We further mitigated this exponential search process by introducing the dynamic transfer network to learn the best transfer learning settings for a given hierarchical architecture.
We showed the generalization of this model across different named entity recognition datasets. For future work, we plan to explore our model on other sequential problems such as  translation, summarization, chat bots as well as explore more advanced gating schemes.

\bibliographystyle{aaai} 
\bibliography{tl.bib}

\begin{thebibliography}{}

\bibitem[\protect\citeauthoryear{Augenstein, Ruder, and
  S{\o}gaard}{2018}]{augenstein2018multi}
Augenstein, I.; Ruder, S.; and S{\o}gaard, A.
\newblock 2018.
\newblock Multi-task learning of pairwise sequence classification tasks over
  disparate label spaces.
\newblock {\em arXiv preprint arXiv:1802.09913}.

\bibitem[\protect\citeauthoryear{Bhatia \bgroup et al\mbox.\egroup
  }{2019}]{bhatia2019comprehend}
Bhatia, P.; Celikkaya, B.; Khalilia, M.; and Senthivel, S.
\newblock 2019.
\newblock Comprehend medical: a named entity recognition and relationship
  extraction web service.
\newblock {\em arXiv preprint arXiv:1910.07419}.

\bibitem[\protect\citeauthoryear{Bhatia, Guthrie, and
  Eisenstein}{2016}]{bhatia2016morphological}
Bhatia, P.; Guthrie, R.; and Eisenstein, J.
\newblock 2016.
\newblock Morphological priors for probabilistic neural word embeddings.
\newblock In {\em Proceedings of the 2016 Conference on Empirical Methods in
  Natural Language Processing},  490--500.

\bibitem[\protect\citeauthoryear{Bojanowski \bgroup et al\mbox.\egroup
  }{2016}]{bojanowski2016enriching}
Bojanowski, P.; Grave, E.; Joulin, A.; and Mikolov, T.
\newblock 2016.
\newblock Enriching word vectors with subword information.
\newblock {\em arXiv preprint arXiv:1607.04606}.

\bibitem[\protect\citeauthoryear{Chen \bgroup et al\mbox.\egroup
  }{2015}]{chen2015mxnet}
Chen, T.; Li, M.; Li, Y.; Lin, M.; Wang, N.; Wang, M.; Xiao, T.; Xu, B.; Zhang,
  C.; and Zhang, Z.
\newblock 2015.
\newblock Mxnet: A flexible and efficient machine learning library for
  heterogeneous distributed systems.
\newblock {\em arXiv preprint arXiv:1512.01274}.

\bibitem[\protect\citeauthoryear{Chiu and Nichols}{2016}]{chiu2016named}
Chiu, J., and Nichols, E.
\newblock 2016.
\newblock Named entity recognition with bidirectional lstm-cnns.
\newblock {\em Transactions of the Association of Computational Linguistics}
  4(1):357--370.

\bibitem[\protect\citeauthoryear{Fan \bgroup et al\mbox.\egroup
  }{2017}]{fan2017transfer}
Fan, X.; Monti, E.; Mathias, L.; and Dreyer, M.
\newblock 2017.
\newblock Transfer learning for neural semantic parsing.
\newblock {\em arXiv preprint arXiv:1706.04326}.

\bibitem[\protect\citeauthoryear{Francis-Landau, Durrett, and
  Klein}{2016}]{francis2016capturing}
Francis-Landau, M.; Durrett, G.; and Klein, D.
\newblock 2016.
\newblock Capturing semantic similarity for entity linking with convolutional
  neural networks.
\newblock {\em arXiv preprint arXiv:1604.00734}.

\bibitem[\protect\citeauthoryear{Graves, Mohamed, and
  Hinton}{2013}]{graves2013speech}
Graves, A.; Mohamed, A.-r.; and Hinton, G.
\newblock 2013.
\newblock Speech recognition with deep recurrent neural networks.
\newblock In {\em Acoustics, speech and signal processing (icassp), 2013 ieee
  international conference on},  6645--6649.
\newblock IEEE.

\bibitem[\protect\citeauthoryear{Guo, Pasunuru, and
  Bansal}{2018a}]{guo2018dynamic}
Guo, H.; Pasunuru, R.; and Bansal, M.
\newblock 2018a.
\newblock Dynamic multi-level multi-task learning for sentence simplification.
\newblock {\em arXiv preprint arXiv:1806.07304}.

\bibitem[\protect\citeauthoryear{Guo, Pasunuru, and
  Bansal}{2018b}]{guo2018soft}
Guo, H.; Pasunuru, R.; and Bansal, M.
\newblock 2018b.
\newblock Soft layer-specific multi-task summarization with entailment and
  question generation.
\newblock {\em arXiv preprint arXiv:1805.11004}.

\bibitem[\protect\citeauthoryear{Hochreiter and
  Schmidhuber}{1997}]{hochreiter1997long}
Hochreiter, S., and Schmidhuber, J.
\newblock 1997.
\newblock Long short-term memory.
\newblock {\em Neural computation} 9(8):1735--1780.

\bibitem[\protect\citeauthoryear{Kingma and Ba}{2014}]{kingma2014adam}
Kingma, D.~P., and Ba, J.
\newblock 2014.
\newblock Adam: A method for stochastic optimization.
\newblock {\em arXiv preprint arXiv:1412.6980}.

\bibitem[\protect\citeauthoryear{Lample \bgroup et al\mbox.\egroup
  }{2016}]{lample2016neural}
Lample, G.; Ballesteros, M.; Subramanian, S.; Kawakami, K.; and Dyer, C.
\newblock 2016.
\newblock Neural architectures for named entity recognition.
\newblock In {\em Proceedings of NAACL-HLT},  260--270.

\bibitem[\protect\citeauthoryear{Liu \bgroup et al\mbox.\egroup
  }{2017}]{liu2017empower}
Liu, L.; Shang, J.; Xu, F.; Ren, X.; Gui, H.; Peng, J.; and Han, J.
\newblock 2017.
\newblock Empower sequence labeling with task-aware neural language model.
\newblock {\em arXiv preprint arXiv:1709.04109}.

\bibitem[\protect\citeauthoryear{McCann \bgroup et al\mbox.\egroup
  }{2018}]{mccann2018natural}
McCann, B.; Keskar, N.~S.; Xiong, C.; and Socher, R.
\newblock 2018.
\newblock The natural language decathlon: Multitask learning as question
  answering.
\newblock {\em arXiv preprint arXiv:1806.08730}.

\bibitem[\protect\citeauthoryear{Mikolov \bgroup et al\mbox.\egroup
  }{2010}]{mikolov2010recurrent}
Mikolov, T.; Karafi{\'a}t, M.; Burget, L.; {\v{C}}ernock{\`y}, J.; and
  Khudanpur, S.
\newblock 2010.
\newblock Recurrent neural network based language model.
\newblock In {\em Eleventh Annual Conference of the International Speech
  Communication Association}.

\bibitem[\protect\citeauthoryear{Peng and Dredze}{2017}]{peng2017multi}
Peng, N., and Dredze, M.
\newblock 2017.
\newblock Multi-task domain adaptation for sequence tagging.
\newblock In {\em Proceedings of the 2nd Workshop on Representation Learning
  for NLP},  91--100.

\bibitem[\protect\citeauthoryear{Pennington, Socher, and
  Manning}{2014}]{pennington2014glove}
Pennington, J.; Socher, R.; and Manning, C.
\newblock 2014.
\newblock Glove: Global vectors for word representation.
\newblock In {\em Proceedings of the 2014 conference on empirical methods in
  natural language processing (EMNLP)},  1532--1543.

\bibitem[\protect\citeauthoryear{Pradhan \bgroup et al\mbox.\egroup
  }{2013}]{pradhan2013towards}
Pradhan, S.; Moschitti, A.; Xue, N.; Ng, H.~T.; Bj{\"o}rkelund, A.; Uryupina,
  O.; Zhang, Y.; and Zhong, Z.
\newblock 2013.
\newblock Towards robust linguistic analysis using ontonotes.
\newblock In {\em Proceedings of the Seventeenth Conference on Computational
  Natural Language Learning},  143--152.

\bibitem[\protect\citeauthoryear{Ratinov and Roth}{2009}]{RatinovRo09}
Ratinov, L., and Roth, D.
\newblock 2009.
\newblock Design challenges and misconceptions in named entity recognition.
\newblock In {\em CoNLL}.

\bibitem[\protect\citeauthoryear{Sachan, Xie, and
  Xing}{2017}]{sachan2017effective}
Sachan, D.~S.; Xie, P.; and Xing, E.~P.
\newblock 2017.
\newblock Effective use of bidirectional language modeling for medical named
  entity recognition.
\newblock {\em arXiv preprint arXiv:1711.07908}.

\bibitem[\protect\citeauthoryear{See, Liu, and Manning}{2017}]{see2017get}
See, A.; Liu, P.~J.; and Manning, C.~D.
\newblock 2017.
\newblock Get to the point: Summarization with pointer-generator networks.
\newblock In {\em Proceedings of the 55th Annual Meeting of the Association for
  Computational Linguistics (Volume 1: Long Papers)}, volume~1,  1073--1083.

\bibitem[\protect\citeauthoryear{Srivastava, Greff, and
  Schmidhuber}{2015}]{srivastava2015training}
Srivastava, R.~K.; Greff, K.; and Schmidhuber, J.
\newblock 2015.
\newblock Training very deep networks.
\newblock In {\em Proceedings of the 28th International Conference on Neural
  Information Processing Systems-Volume 2},  2377--2385.
\newblock MIT Press.

\bibitem[\protect\citeauthoryear{Strubell \bgroup et al\mbox.\egroup
  }{2017}]{strubell2017fast}
Strubell, E.; Verga, P.; Belanger, D.; and McCallum, A.
\newblock 2017.
\newblock Fast and accurate entity recognition with iterated dilated
  convolutions.
\newblock In {\em Proceedings of the 2017 Conference on Empirical Methods in
  Natural Language Processing},  2670--2680.

\bibitem[\protect\citeauthoryear{Tjong Kim~Sang and
  De~Meulder}{2003}]{tjong2003introduction}
Tjong Kim~Sang, E.~F., and De~Meulder, F.
\newblock 2003.
\newblock Introduction to the conll-2003 shared task: Language-independent
  named entity recognition.
\newblock In {\em Proceedings of the seventh conference on Natural language
  learning at HLT-NAACL 2003-Volume 4},  142--147.
\newblock Association for Computational Linguistics.

\bibitem[\protect\citeauthoryear{Uzuner, Solti, and
  Cadag}{2010}]{uzuner2010extracting}
Uzuner, {\"O}.; Solti, I.; and Cadag, E.
\newblock 2010.
\newblock Extracting medication information from clinical text.
\newblock {\em Journal of the American Medical Informatics Association}
  17(5):514--518.

\bibitem[\protect\citeauthoryear{Verga, Strubell, and
  McCallum}{2018}]{verga2018simultaneously}
Verga, P.; Strubell, E.; and McCallum, A.
\newblock 2018.
\newblock Simultaneously self-attending to all mentions for full-abstract
  biological relation extraction.
\newblock {\em arXiv preprint arXiv:1802.10569}.

\bibitem[\protect\citeauthoryear{Vinyals, Fortunato, and
  Jaitly}{2015}]{vinyals2015pointer}
Vinyals, O.; Fortunato, M.; and Jaitly, N.
\newblock 2015.
\newblock Pointer networks.
\newblock In {\em Advances in Neural Information Processing Systems},
  2692--2700.

\bibitem[\protect\citeauthoryear{Wang \bgroup et al\mbox.\egroup
  }{2018}]{wang2018label}
Wang, Z.; Qu, Y.; Chen, L.; Shen, J.; Zhang, W.; Zhang, S.; Gao, Y.; Gu, G.;
  Chen, K.; and Yu, Y.
\newblock 2018.
\newblock Label-aware double transfer learning for cross-specialty medical
  named entity recognition.
\newblock {\em arXiv preprint arXiv:1804.09021}.

\bibitem[\protect\citeauthoryear{Williams and
  Zipser}{1989}]{williams1989learning}
Williams, R.~J., and Zipser, D.
\newblock 1989.
\newblock A learning algorithm for continually running fully recurrent neural
  networks.
\newblock {\em Neural computation} 1(2):270--280.

\bibitem[\protect\citeauthoryear{Yang, Salakhutdinov, and
  Cohen}{2016}]{yang2016multi}
Yang, Z.; Salakhutdinov, R.; and Cohen, W.
\newblock 2016.
\newblock Multi-task cross-lingual sequence tagging from scratch.
\newblock {\em arXiv preprint arXiv:1603.06270}.

\bibitem[\protect\citeauthoryear{Yang, Salakhutdinov, and
  Cohen}{2017}]{yang2017transfer}
Yang, Z.; Salakhutdinov, R.; and Cohen, W.~W.
\newblock 2017.
\newblock Transfer learning for sequence tagging with hierarchical recurrent
  networks.
\newblock {\em arXiv preprint arXiv:1703.06345}.

\end{thebibliography}
\end{document}